\documentclass[journal]{IEEEtran}

\usepackage{mathtools}
\usepackage{makecell}
\usepackage{threeparttable}
\usepackage{caption}
\usepackage{array}
\usepackage{booktabs}
\usepackage{multirow}
\usepackage[bookmarks=false]{hyperref}
\usepackage{pifont}
\usepackage{cite}
\usepackage{amsmath,amsfonts}
\usepackage{algorithmic}
\usepackage{graphicx}
\usepackage{textcomp}

\usepackage{amssymb}
\usepackage{latexsym}

\usepackage{url}
\usepackage[dvipsnames]{xcolor}
\definecolor{mypink1}{rgb}{0.858, 0.188, 0.478}
\usepackage{soul} 

\begin{document}

\title{SRPN: similarity-based region proposal networks for nuclei and cells detection in histology images}

\author{Yibao Sun${}^{1}$, Xingru Huang${}^{1\star}$, Huiyu Zhou${}^{2}$, Qianni Zhang${}^{1}$
\thanks{${}^\star$Corresponding author: Xingru Huang.}
\thanks{${}^{1}$Yibao Sun, Xingru Huang and Qianni Zhang are with the School of Electronic Engineering and Computer Science, Queen Mary University of London, Mile End Road, London, E1 4NS, United Kingdom (e-mail: yibao.sun; xingru.huang; qianni.zhang@qmul.ac.uk).}
\thanks{${}^{2}$Huiyu Zhou is with the School of Informatics, University of Leicester, University Road, Leicester, LE1 7RH, United Kingdom (e-mail: hz143@leicester.ac.uk).}
}

\maketitle

\begin{abstract}
The detection of nuclei and cells in histology images is of great value in both clinical practice and pathological studies. However, multiple reasons such as morphological variations of nuclei or cells make it a challenging task where conventional object detection methods cannot obtain satisfactory performance in many cases. A detection task consists of two sub-tasks, classification and localization. Under the condition of dense object detection, classification is a key to boost the detection performance. Considering this, we propose similarity based region proposal networks (SRPN) for nuclei and cells detection in histology images. In particular, a customized convolution layer termed as embedding layer is designed for network building. The embedding layer is added into the region proposal networks, enabling the networks to learn discriminative features based on similarity learning. Features obtained by similarity learning can significantly boost the classification performance compared to conventional methods. SRPN can be easily integrated into standard convolutional neural networks architectures such as the Faster R-CNN and RetinaNet. We test the proposed approach on tasks of multi-organ nuclei detection and signet ring cells detection in histological images. Experimental results show that networks applying similarity learning achieved superior performance on both tasks when compared to their counterparts. In particular, the proposed SRPN achieve state-of-the-art performance on the MoNuSeg benchmark for nuclei segmentation and detection while compared to previous methods, and on the signet ring cell detection benchmark when compared with baselines. The sourcecode is publicly available at: \textcolor{mypink1}{\url{https://github.com/sigma10010/nuclei_cells_det}}.
\end{abstract}


%
\IEEEpeerreviewmaketitle

\section{Introduction}
Pathology has benefited from the rapid progress in technology of digital scanning during the last decade. Nowadays, slide scanners are able to produce super-resolution whole slide images (WSI) \cite{gilbertson2005clinical}, also called digital slides, which can be explored by image viewers as an alternative to the use of conventional microscope. The use of WSI together with the other microscopic and molecular pathology images brings the development of digital pathology, which further enables to perform digital diagnostics. Standardization efforts of digital pathology has been made in Europe \cite{rojo2012standardization}. Moreover, the availability of WSI makes it possible to apply image processing and recognition techniques to support digital diagnostics, opening new revenues of computational pathology. There have been some computational pathology tools that support pathologists for very routine tasks such as to segment nuclei \cite{song2017dual,graham2019hover,zhao2020triple} or tumour \cite{qaiser2019fast} and to classify cancer in histopathological images \cite{xu2014weakly,reis2017automated,li2019analysis}. Due to the promising impact on future pathology practice,  digital pathology and computational pathology have been attracting tremendous attention \cite{al2012digital,louis2015computational}.

Cancer diagnosis and prognosis based on digital slides is of significant value both in clinical medicine and pathological research. A pathology report that gives detailed information on the assessment of cancer stage and progression can help employ personalised therapy and provide better treatment and care post tumour resection surgery. Generally, cancer staging is determined by various aspects such as differentiation of tissues, morphological variety and distribution of cells. In a routine of cancer staging, pathologists need to frequently perform several necessary operations to examine digital slides, such as identifying certain cells or nuclei, marking them or counting them. The procedure is labor-intensive and often leads to inter-observer disagreement. Well-trained specialists often report different opinions against each other. According to the definition given in \cite{louis2014computational}, computational pathology is a promising solution to improve pathological routine efficiency and to eliminate inter-observer variability. However, training more effective computational algorithms requires adequate data and obtaining large-scale annotated pathology datasets by pathologists is expensive. 
Even when adequate annotated pathology datasets are available, the intrinsic complex morphological characteristics and variations keep histology image analysis a  challenging task. 

In recent years, benefiting from the powerful computational resources and the availability of large-scale labeled data, deep learning has made incredible advances in image recognition related challenges, and has become a solution for computational pathology. 
In many cases, morphological and numeric features of nuclei and cells are meaningful for cancer assessment. For instance, the Nottingham system grades breast cancer by adding up scores for tubule formation, nuclear pleomorphism and mitotic count \cite{ellis1991pathologic}. Among these factors, nuclear pleomorphism could give an indication of the degree of the cancer evolution while mitotic count could give an evaluation of the aggressiveness of the tumour. Cell-level analysis is normally performed by pathologists manually by using a microscope or examining digital slides. This process is laborious, error-prone and sometimes impossible due to the high density of cell in some regions. Thus, it is highly demanding to build a computational model that is able to automatically and accurately detect, segment and quantify nuclei and cells of interest in a digital slide. 

\begin{figure}
\centering
\includegraphics[width=0.45\textwidth]{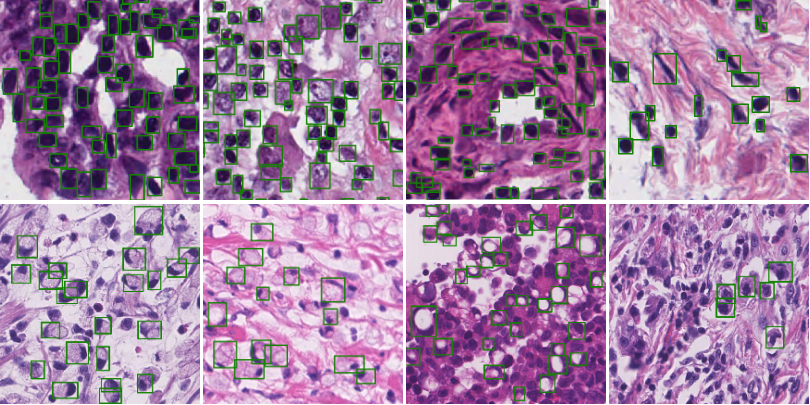}
\caption{\label{fig:nucleicell} Variations of nuclei (first row) and signet ring cells (second row) in histology images.}
\end{figure}

Histology images produced by different laboratories with different platforms unavoidably introduce variations in colour, scale and shape of nuclei and cells (Fig. \ref{fig:nucleicell}). Overlapping cells poses further intrinsic complications to the task. There are also some external factors that add difficulties to the cell detection task, e.g., the lack of quality and quantity in the annotation labels and class imbalance, which impose widely encountered and long lasting issues in biomedical image analysis. Various CNN based systems have been developed to resolve the task of cell detection. Some works directly apply well-developed object detectors of excellent performance on cell detection. For example, Zhang et al. \cite{zhang2016cancer} successfully apply the framework of Faster R-CNN \cite{ren2015faster} to detect adhesion cells in phase-contrast microscopy images; Yi et al. \cite{yi2017fast} solve the task of accurate neural cell detection by adapting the original SSD to a light-weight model. Although those deep learning based systems succeed in some specific cases, they cannot obtain satisfactory performance in more general scenarios. 

The heterogeneity in cell-level objects and the visual challenges existing in histology images together make the classification, detection and segmentation of these objects a completely different task than working on objects in natural images. The unique morphological nature of cells and nuclei need to be considered and specifically addressed in the design of relevant deep learning solutions. Thus, in this research a dedicated similarity learning enhanced deep neural network is presented with the leverage of state-of-the-art techniques to detect generic cell-level objects in histology images.
The main contributions include: 1) Tailored similarity-based region proposal networks for solving the challenges in nuclei and cells detection in histology images, with special focus on detecting individual nuclei instances in cases where high visual variance and intense occlusion take place. 2) A new network architecture that includes embedding layers to enable similarity learning, providing expressive and discriminative features that suit the task of nuclei and cells detection. 3) The proposed method is applied in solving two different tasks - multi-organ nuclei detection and signet ring cell detection - to validate the effectiveness of the proposed method compared against the state-of-the-arts. Multiple CNN architectures are tested to reveal their impacts on nuclei or cells detection. Different loss functions are applied to the training of the networks. 

\section{Related work}

\subsection{Object detection}
\label{sec: od}
Visual object detection is defined as localising and categorising objects of interest in a given image. Classical framework of detectors mainly consist of three processes: 1) propose regions of interest (ROI) to predict candidate bounding box; 2) extract feature vectors from ROI for classification; 3) categorize ROI and refine the corresponding bounding boxes. Generally, a sliding window approach is used to search for ROI. To better consider situations where objects entail scale and aspect ratio variations, some strategies have been proposed such as cropping the input image into different sizes or using multiple sliding windows with different aspect ratios \cite{vedaldi2009multiple, viola2001rapid}.

R-CNN is a pioneering framework that exploits regional features extracted by CNN for object detection \cite{girshick2014rich}. Compared to the previous complex ensemble systems like SegDPM \cite{fidler2013bottom}, R-CNN makes a breakthrough and achieved a mean average precision (mAP) of 53.3\% on the detection benchmark VOC 2012. However, each module of R-CNN must be trained separately, making it difficult to obtain a global optimisation. Fast R-CNN is proposed to address this limitation of R-CNN \cite{girshick2015fast}. The training of Fast R-CNN is performed in an end-to-end manner by using a multi-task loss. Moreover, Fast R-CNN introduces a ROI Pooling layer to extract regional features from feature maps. The ROI Pooling layer applies max pooling to convert features inside each reasonable ROI into a small uniform feature map. These changes make Fast R-CNN a better detector both in accuracy and inference speed compared to R-CNN. Still, the conventional region proposal methods used by Fast R-CNN are computationally expensive and based on hand-crafted features, which poses limitations on the performance. To eliminate these limitations of  region based detectors, Ren et al.  proposed region proposal networks (RPN) \cite{ren2015faster}, devising a data-driven and learnable way for region proposals. The resulting detector Faster R-CNN demonstrates an outstanding performance with a very high inference speed, making it a real-time object detection system.

Liu et al. propose to use feature pyramid networks (FPN) to solve the scale variation problem faced by object detection \cite{lin2017feature}. The network architecture of FPN is similar to the one used in U-net \cite{ronneberger2015u} and stacked hourglass networks \cite{newell2016stacked}. Applying FPN in adapted single-scale detectors like RPN, Fast R-CNN and Faster R-CNN leads to significant improvement on detection accuracy for each baseline without increasing the inference time. The adapted Faster R-CNN reported state-of-the-art results on the COCO detection benchmark \cite{lin2014microsoft}.

Besides scale variation, class imbalance between background and foreground is another challenge in object detection. Instead of using the strategy of hard example mining, which intuitively discard some easy negative examples and sample a fixed ratio (e.g., 3:1) between negatives and positives  \cite{shrivastava2016training,bucher2016hard}, Lin et al. introduce a novel focal loss to address the problem of class imbalance by suppressing the gradients of easy samples \cite{lin2017focal}. The resulting detector named RetinaNet is built on the basis of RPN and FPN, but trained by the focal loss which is able to match the speed of previous one-stage detectors while surpassing the accuracy of all existing state-of-the-art two-stage detectors.

\subsection{Similarity learning}
Similarity learning is a promising way to learn effective visual representations without human supervision \cite{hadsell2006dimensionality, wu2018unsupervised, NEURIPS2019_ddf35421}. These approaches learn visual representations by contrasting positive samples against negative samples. To learn from unlabeled data, in \cite{dosovitskiy2014discriminative}, the authors propose to treat each instance as a class and perform a variety of transformations to each instance to yield training sets with surrogate labels. Through using the instance classes we can discard human supervision. Meanwhile, large computational complexity imposed by learning from instance classes becomes a new challenge. The memory bank has been proposed to tackle the computational problem \cite{wu2018unsupervised, he2020momentum, misra2020self}. Instead of using a memory bank, some works use in-batch samples for negative sampling \cite{doersch2017multi,ye2019unsupervised,ji2019invariant}. With pairing samples, distance metrics in an embedding space are used to measure the similarity between samples. Similar samples are closer than those dissimilar ones in the embedding space. Various loss functions based on distance metric in an embedding space, such as the contrastive loss \cite{hadsell2006dimensionality} and the triplet loss \cite{hoffer2015deep}, have been proposed for similarity learning. Similarity learning becomes widely used for tasks like signature verification \cite{bromley1994signature}, one-shot image recognition \cite{koch2015siamese} and object tracking \cite{bertinetto2016fully}. Despite of its success, similarity learning is rarely used in analysing histopathological images. In this paper, we show the effectiveness of learning multi-scale embeddings in a contrastive way for nuclei and cell detection in digital histology images, with an appropriate design.

\subsection{Nuclei and cells detection}
The detection of nuclei and cells is a critical step for cell-level analysis in digitised WSIs, from which useful clinical clues including cell distribution and categorisation can be automatically acquired. Similar to object detection introduced in Section \ref{sec: od}, approaches for cell detection have evolved from using handcrafted features to exploiting learned features. Most early stage approaches exploit handcrafted low-level visual features that encode information such as shape \cite{cheng2008segmentation}, edge \cite{jung2010segmenting}, luminance \cite{faustino2011graph} and texture \cite{irshad2013automated}, to detect nuclei or cells in WSIs.

Nowadays, convolution neural networks are generally recognised to be more powerful to learn image representations from pixel intensity. Due to its superior ability in learning robust features, a variety of works employ CNNs to tackle the task of cell detection. A straightforward way to use classification networks for detection related tasks is to train a classifier with small image patches for target objects, and then apply the trained classifier to make predictions on a large input image with the help of a sliding-window, whose center pixel is classified as background or foreground. Some early works following this approach show promising results for the purpose of cell-level object detection, including mitosis detection \cite{cirecsan2013mitosis,li2018deepmitosis} and nuclei detection in colon cancer histology images \cite{sirinukunwattana2016locality}. However, a major drawback of this kind of methods is that they are unable to deal with object scale variation. Instead of using classification architectures for nuclei and cell detection, further works deploy regional CNN (R-CNN) architectures where the scale variation problem is well considered. For instance, Xu et al. integrate an improved U-net and SSD to detect and segment cell instances in a multi-task way \cite{xu2019us}. In practice however, the CNNs designed and trained for natural images are often unable to achieve satisfactory performance when directly applied to biomedical images.

An alternative method to detect a nucleus is to localize its center, instead of using a bounding box. Several works study nuclei detection in this setting \cite{kainz2015you,zhou2018sfcn}. In \cite{kainz2015you}, a regression model predicts and outputs a score map of the same size as the input image. Each pixel value of the score map indicates its inverted distance to the nearest nucleus center. Local extremums of the score map are then considered as nuclei centers. The model is simple and easy to implement, but its performance relies on cell density and a hypothesis that all nuclei are in a circle shape.

Nuclei segmentation is another area that attracts significant attention. The multi-organ nuclei segmentation (MoNuSeg) dataset, which is used for testing nuclei detection methods in this paper, also supports nuclei segmentation \cite{kumar2017dataset}. Based on this dataset, several nuclei segmentation solutions are presented \cite{yoo2019pseudoedgenet,graham2019hover,zhao2020triple}. These approaches are mainly base on U-Net \cite{ronneberger2015u}, with auxiliary strategies like nuclear contour regularization \cite{zhou2019cia,wang2020bending} and/or multi-scale feature aggregation \cite{hu2019mc,zhao2020triple}. Zhou et al., propose a contour-aware information aggregation method for nuclei instance segmentation \cite{zhou2019cia}. In their study, besides employing the standard Average Jaccard Index (AJI) for segmentation performance evaluation, they also report the state-of-the-art $F1$-score for nuclei instance detection on the MoNuSeg dataset. Comparison between the proposed SRPN method against these previous approaches based on valid metrics are presented in the Experiments section \ref{sec: ex}.

\section{Methodology}
\begin{figure}
\centering
\includegraphics[width=0.3\textwidth]{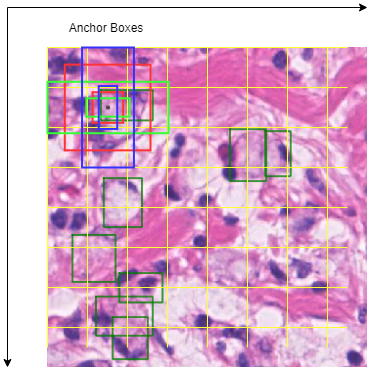}
\caption{\label{fig:anchor} Illustration of using anchor boxes for object detection. For each location in the feature map, multiple anchor boxes with different scales and aspect ratios are considered as candidates. The yellow grid roughly denotes the receptive field of neural networks.}
\end{figure}

Given an image, one common method to detect objects of interest across the whole image is to use anchor boxes \cite{ren2015faster}. As illustrated in Fig. \ref{fig:anchor}, at first a large number of anchor boxes (object bounding boxes) serving as object (cell) candidates are overlaid on each possible locations of the input image. Network (detector) parameters are then adjusted to simultaneously refine the candidate bounding boxes and to assign a label for each candidate bounding box during the process of training. Normally, to take into account the difference in size and shape of object, multiple anchor boxes with different scales and aspect ratios are assigned for each candidate location. In our experiments, we use 3 scales and 3 aspect ratios, 9 anchor boxes per location. There are many methods to adjust parameters of a detector. The proposed method exploits the advantage of similarity learning to achieve a high performance for cell-level object detection. Next, we describe the proposed method in detail focusing on two aspects, i.e., the network architecture and loss functions. 

\begin{figure*}
\centering
\includegraphics[width=1.0\textwidth]{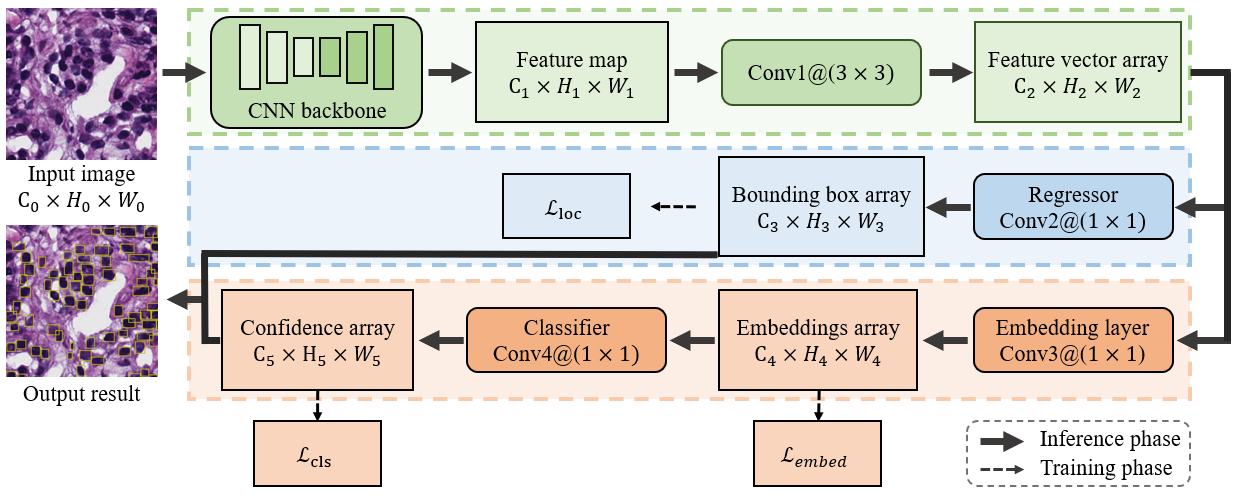}
\caption{\label{fig:ringcelldetect} Architecture of SRPN for nuclei and cells detection. It takes an image as input and outputs prediction indicating locations and confidences for nuclei across whole image by using a regressor and classifier head respectively. 
}
\end{figure*}

\subsection{Network architecture}
\label{sec:net}
The proposed network architecture to detect nuclei and cells is illustrated in Fig. \ref{fig:ringcelldetect}. At the beginning, a CNN backbone is used to extract feature maps from an input image of size $C_{0}\times H_{0}\times W_{0}$ ($C_{0}=3$ for RGB image), since features extracted by CNN have been demonstrated with excellent robustness to various kinds of visually related tasks such as classification \cite{krizhevsky2012imagenet}, segmentation \cite{long2015fully} and detection \cite{girshick2014rich}.
Given an extracted feature map of $C_1$ channels as input, a convolution layer (Conv1) encodes a local region of $3 \times 3$ pixels of the feature map into a feature vector of length $C_{2}$; in our experiments $C_1=C_2=256$.
Predictions of a bounding box array and a confidence array are then acquired for each feature vector ($H_{2} \times W_{2}$ in total) by using a regressor and classifier head respectively. A regressor head (Conv2) encodes offsets between the default anchor boxes and the corresponding predicted bounding boxes. A classifier head (Conv4) assigns a confidence score indicating foreground or not for each predicted bounding box with a Softmax function. Before the classifier head, an embedding layer (Conv3) is added to enable similarity learning to improve the classification performance. In order to keep location consistency, a convolution layer with a kernel size of $1 \times 1$ is applied to the regressor, the classifier or the embedding layer. An anchor box is presented as a 4-tuple, consisting of a coordinate pair of its top left corner and its height and width, that is to say, $C_{3} = 4\times num\_anchor$ where $num\_anchor$ denotes the number of anchors to predict per location. $C_{4} = num\_anchor \times dim\_embedding$, $C_{5} = num\_anchor$, where $dim\_embedding$ denotes the dimension of embedding, set to 20 in our experiments.

In contrast to the original RPN settings proposed by \cite{ren2015faster}, an embedding layer is added before the classifier head to enable similarity learning with the aim to improve the performance of nuclei detection. The motivations behind applying similarity learning in this framework are in two folds. On the one hand, embeddings learned under the constraint that samples of the same class are clustering and those of different classes are separating, are more discriminative, especially in the cases of identifying one specific type of objects out of a noisy background. A well-performing classifier is of crucial importance to build an excellent object detector. On the other hand, through pairing samples for similarity learning, one can indirectly eliminate the impact of the class imbalance problem commonly faced by object detectors by controlling the sampling process. Furthermore, we can generate the maximum of $n^2$ sample pairs or $n^3$ sample triplets from $n$ training samples, meaning that the pairing of samples also serves as a data augmentation process for model training. Overall, the similarity learning paradigm demonstrates significant benefits for feature learning in object detection tasks.

\subsection{Loss functions}
\label{sec:loss_f}
According to the network architecture presented in Section \ref{sec:net}, when given an image with ground truth, the embedding layer outputs an embedding array of size $C_4 \times H_4 \times W_4$ where $C_4$ equals to the product of the number of the anchors per location and the dimension of the embeddings, i.e., $9 \times 20$ in our experiments. To perform supervised learning, a label indicating foreground or background is assigned to each anchor based on the intersection over union (IoU) between the anchor and the corresponding ground truth. An anchor is given a positive label $1$ if it has an IoU higher than the positive threshold, say 0.7, with any ground truth box. A negative label $0$ is given to an anchor if the IoU is lower than the negative threshold, say 0.3, with all the ground truth boxes. Anchors that are neither positive nor negative will be filtered out during training. 

To use similarity learning, generating embedding pairs or triplets is a key step. Given a set of embeddings $\mathcal{E}_{1} = \left \{ \left ( \mathbf{\epsilon}_{i}, p_{i}^{*} \right ) | i\in \mathbb{Z^{+}} \right \}$, where $\epsilon_i$ represents embedding for the $i^{th}$ anchor and $p_i^* \in \{0,1\}$ denotes its anchor label, it is easy to transform $\mathcal{E}_{1}$ into 1) a set of embedding pairs $\mathcal{E}_{2} = \left \{ \left ( \mathbf{\epsilon}_{i}, \mathbf{\epsilon}_{i}', s_{i} \right ) | i \in \mathbb{Z}^{+}\right \}$, where $s_{i} \in \left \{ 0,1 \right \}$ denotes the similarity/closeness between embedding $\mathbf{\epsilon}_{i}$ and $\mathbf{\epsilon}_{i}'$; or 2) a set of embedding triplets $\mathcal{E}_{3} = \left \{ \left ( \mathbf{\epsilon}_{i}^{a}, \mathbf{\epsilon}_{i}^{p}, \mathbf{\epsilon}_{i}^{n} \right ) | i \in \mathbb{Z}^{+}\right \}$, where $\mathbf{\epsilon}_{i}^{a}$ is a reference embedding, and $\mathbf{\epsilon}_{i}^{p}$ is a positive embedding of the same class as the reference while $\mathbf{\epsilon}_{i}^{n}$ is a negative embedding of a different class. In practice, the sampling process can be controlled to the balance embedding pairs with a different label $s_i$. For a better description, we define a function $\varphi: \mathcal{E}_{1} \to \mathcal{E}_{2}$ to represent the process of generating pairs, and $\psi : \mathcal{E}_{1} \to \mathcal{E}_{3}$ to represent the generation of triplets.

With the embedding pairs $\mathcal{E}_{2}$ or triplets $\mathcal{E}_{3}$, we can now apply the contrastive loss/pair loss \cite{hadsell2006dimensionality} or triplet loss \cite{hoffer2015deep} as a constraint for similarity learning. The pair loss is defined as follows:

\begin{equation}
\mathbf{\mathcal{L}}_{pair}(\mathbf{\epsilon,\epsilon'},s) = \frac{1}{2}s\left \| \mathbf{\epsilon}-\mathbf{\epsilon'} \right \|^{2}+ \frac{1}{2}(1-s) max(m-\left \| \mathbf{\epsilon}-\mathbf{\epsilon'} \right \|^{2}, 0)
\label{eq: contrastive}
\end{equation}
where $m$ is a constant of margin, and $\left \| \cdot  \right \|$ an Euclidean distance metric. After the process of minimizing the loss function, the distance between two samples with different categories should be greater than the margin $m$. In other words, samples of different classes spread widely in the embedding space. Meanwhile, samples of the same class cluster closely together. Embeddings learned in this way are expected to be capable of discriminating sample classes.

The triplet loss is defined as:
\begin{equation}
\begin{aligned}
\mathbf{\mathcal{L}}_{triplet}(\mathbf{\epsilon^{a},\epsilon^{p},\epsilon^{n}}) = max(\left \| \mathbf{\epsilon^{a}}-\mathbf{\epsilon^{p}} \right \|^{2}-\left \| \mathbf{\epsilon^{a}}-\mathbf{\epsilon^{n}} \right \|^{2}+m , 0)
\label{eq: triplet}
\end{aligned}
\end{equation}
where $m$ and $\left \| \cdot  \right \|$ denote the same as in the pair loss. After optimisation, the distance between a positive pair should be less than that between a negative pair by a margin $m$.

Following the R-CNN based approaches for object detection \cite{girshick2014rich,girshick2015fast,ren2015faster}, a classification head and a regression head are employed for object identification and bounding box regression respectively, as depicted in Fig. \ref{fig:ringcelldetect}. For the classification head, a regular cross-entropy loss or focal loss \cite{lin2017focal} is used for weights tuning. For the regression head, following \cite{girshick2015fast}, we apply the smooth $L_{1}$ loss for anchor box tuning. 
An anchor box is encoded as a 4-tuple $[x_{a}, y_{a}, h_{a}, w_{a}]$, where ($x_{a}, y_{a}$) indicate the coordinate of its top left corner, and  $(h_{a}, w_{a})$ represent its height and width respectively. To refine the anchor boxes, offsets between the final predicted bounding box and the corresponding anchor box are encoded as a 4-tuple $\mathbf{t} = [t_{x}, t_{y}, t_{h}, t_{w}]$ such that:
\begin{equation}
\begin{bmatrix*}[l]
t_x=\left( x-x_a \right )/w_a \\ 
t_y=\left( y-y_a \right )/h_a \\ 
t_h=log\left( h/h_a \right ) \\ 
t_w=log\left( w/w_a \right )
\end{bmatrix*}
\label{eq: reg_off}
\end{equation}
where $[x, y, h, w]$ is the 4-tuple for the final predicted bounding box similar to $[x_{a}, y_{a}, h_{a}, w_{a}]$ for the anchor box. In a supervised learning setting, ground truth bounding boxes are also input as supervision signals. The offsets between a ground truth bounding box and an anchor box are encoded as $\mathbf{t}^{*} = [t_{x}^{*}, t_{y}^{*}, t_{h}^{*}, t_{w}^{*}]$, such that:
\begin{equation}
\begin{bmatrix*}[l]
t_x^*=\left ( x^*-x_a \right )/w_a \\ 
t_y^*=\left ( y^*-y_a \right )/h_a \\ 
t_h^*=log\left ( h^*/h_a \right ) \\ 
t_w^*=log\left ( w^*/w_a \right )
\end{bmatrix*}
\label{eq: reg_off_gt}
\end{equation}
where $[x^{*}, y^{*}, h^{*}, w^{*}]$ is a 4-tuple for a ground truth box.
With the definitions above, the smooth $L_1$ loss can be defined as:
\begin{equation}
\begin{aligned}
\mathcal{L}_{smoothL_{1}}(\mathbf{t}, \mathbf{t}^{*}) &= \sum_{j\in \left \{ x, y, h, w \right \}} f (t_{j}-t_{j}^{*})
\label{eq: reg}
\end{aligned}
\end{equation}
where $f\left ( \cdot  \right )$ is the smooth $L_{1}$ function:
\begin{equation}
\begin{aligned}
f(x)=\begin{cases}
0.5x^{2}& \text{ if } \left |  x \right |< 1 \\ 
 \left |  x \right |-0.5& \text{ otherwise. }
\end{cases}
\end{aligned}
\end{equation}

Overall, the total loss for an input image with the ground truth is a weighted sum of the embedding loss $\mathcal{L}_{embed}$, the localization loss $\mathcal{L}_{loc}$ and the classification loss $\mathcal{L}_{cls}$:
\begin{equation}
\begin{aligned}
\mathcal{L}= \ &\sum_{i}^{N} \mathcal{L}_{embed}\left ( \varepsilon \left (  \mathbf{\epsilon}_{i}, p_{i}^{*} \right ) \right )+ \\
&\sum_{i}^{N}p_{i}^{*}\mathcal{L}_{loc}\left (  \mathbf{t}_{i} ,  \mathbf{t}_{i}^{*}  \right )+\sum_{i}^{N}\mathcal{L}_{cls}\left (  p_{i},p_{i}^{*} \right )
\label{eq: loss}
\end{aligned}
\end{equation}
in which $N$ denotes the number of anchor boxes, $\varepsilon\left ( \cdot  \right )=  \varphi\left ( \cdot  \right )\text{or} \ \psi\left ( \cdot  \right ) $ depending on the option of the embedding loss $\mathcal{L}_{embed}$. In our experiments, we employ either the pair loss in Eq. (\ref{eq: contrastive}) or the triplet loss in Eq. (\ref{eq: triplet}) as the embedding loss $\mathcal{L}_{embed}$. The term $p_{i}^{*}\mathcal{L}_{loc}$ indicates that the localization loss is activated only for positive anchors where $p_{i}^{*}=1$ and is disabled otherwise, $p_{i}^{*}=0$. The smooth $L_{1}$ loss Eq. (\ref{eq: reg}) is tested as the localization loss. Since there is only one cell or nuclei type to detect, as mentioned before, we 
employ either the cross-entropy loss or the focal loss \cite{lin2017focal} as the classification loss.

\subsection{Enhanced Faster R-CNN and RetinaNet}
As described in Section \ref{sec: od}, both Faster R-CNN \cite{ren2015faster} and RetinaNet \cite{lin2017focal} utilise RPN to propose possible foreground regions. It is easy to replace the RPN module with the proposed SRPN so that similarity learning is enabled in the framework to improve both Faster R-CNN and RetinaNet for nuclei and cells detection in histology images.

\section{Experiments}
\label{sec: ex}

\subsection{Training and inference}
The detectors are trained using the optimiser of stochastic gradient descent (SGD) together with a basic learning rate of 1e-3. We validate several CNN architectures, like ResNet-50/ResNet-101 \cite{he2016deep} and ResNeXt-50/ResNeXt-101 \cite{xie2017aggregated},  as the backbone of detectors. To speed up the training procedure, we exploit networks pretrained on ImageNet \cite{deng2009imagenet}. The weights and biases in the other layers are initialized by values drawn from the normal distribution $N\sim( 0,0.01^{2} )$ and a constant of 0 respectively. The training procedure takes a couple of hours for each detector on a GPU of NVIDIA GeForce GTX Titan X, depending on the range of the batch size from 4 to 12.

To ensure the robustness of the detectors against visual variations in histology images, data augmentation is performed during training by transforming the training images in ways of colour jitter, horizontal flipping and vertical flipping with a certain probability of 0.5. For colour jitter, the image colour are randomly changed in their brightness, contrast, saturation and hue. 

Class imbalance is a very common problem faced by dense object detection. Generally, the number of interested objects is much less than the rest of searched locations to predict in an input image. That is to say, the number negative samples overwhelms the number of positive samples. Thus, during the training process, the technique of online hard example mining (OHEM) is applied to eliminate the effect of class imbalance \cite{shrivastava2016training}. Consequently, the ratio between the negative and positive samples becomes $3:1$, similar to that reported in previous works\cite{ren2015faster}.

In the inference phase, there might be several predictions for one object due to the settings designed for dense object detection. Normally, a process of non-maximum suppression is performed to remove the repeated predictions, keeping only one with the highest probability for each object \cite{neubeck2006efficient}. A threshold of $IoU$ between two predictions is used to decide whether they are repeated or not. In our experiments, the threshold is set to 0.3.

\subsection{Evaluation on multi-organ nuclei detection}
Nuclei detection in histology images  enables the extraction of cell-level features for computational histopathology analysis. Once accurately detected, nuclear morphometric and appearance features such as nuclei density, average size, and pleomorphism can be used to assess cancer grades, as well as to predict treatment effectiveness. Identifying different types of nuclei based on the detection results can also yield information about tumour growth, which is important for cancer grading. In this section, we utilise the MoNuSeg dataset \cite{kumar2017dataset} to validate the effectiveness of the proposed method for nuclei detection in histology images.

\subsubsection{Dataset}
The MoNuSeg dataset is published for the Multi-organ Nuclei Segmentation challenge\footnote{\href{https://monuseg.grand-challenge.org/Data/}{https://monuseg.grand-challenge.org/Data/}} in MICCAI 2018. The training dataset consists of 30 images generated from multiple organs including breast, kidney, liver, prostate, bladder, colon and stomach, each of size $1,000 \times 1,000$ pixels. There are in total 21,713 nuclear boundary annotations drawn by domain experts. The testing dataset consists of 14 images with 6,697 additional nuclear boundary annotations. To validate the proposed method thoroughly, the testing dataset is organised into two groups based on the organ types. Images in group 1 are taken from the same organs of training data (seen) and images in group 2 from unseen organs (unseen).
\subsubsection{Evaluation criteria}
The first metric used to validate the effectiveness of the proposed method for nuclei detection is the $F1$-score ($F1=\frac{2TP}{2TP+FP+FN}$). The value of true positives ($TP$) is the number of ground truth objects with a matched predicted object.
The value of false positives ($FP$) is the number of predicted objects without a matched ground truth object.
The value of false negatives ($FN$) equals to the number of the ground truth objects without a matched predicted object. Intersection over union ($IoU$) is computed to decide if two objects are matched or not. In our experiments, the $IoU$ threshold is set as 0.3.
Besides $F1$-score, we also report the average precision ($AP$) value for each test to provide additional evaluation.

\begin{figure}
\centering
\includegraphics[width=0.38\textwidth]{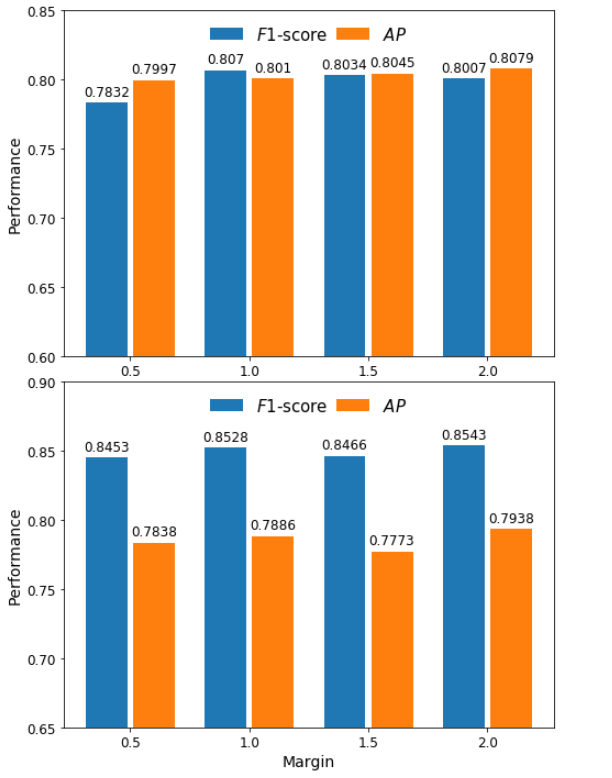}
\caption{\label{fig:sel_m} $F1$-score and average precision ($AP$) against different margins for the pair loss (top one) and the triplet loss (button one) respectively.}
\end{figure}

\begin{table*}[]
\centering
\begin{threeparttable}
\caption{Comparison of the proposed SRPN method with different CNN backbones and embedding loss functions.}
\label{tab:monu_result_archi}
\begin{tabular}{l|l|l|l|l|l|l|l|l|l}
\Xhline{3\arrayrulewidth}
 Model & Backbone & $\mathcal{L}_{embed}$ & $TP$ & $FP$ & $FN$ & Precision & Recall & $F1$-score & $AP$\\
 \Xhline{2\arrayrulewidth}
RPN & ResNet-50 & - & 4760 & 1040 & 1937 & 0.8207 & 0.7108 & 0.7618 & 0.7253 \\
SRPN & ResNet-50 & $\mathcal{L}_{pair}$ & 5131 & 1000 & 1566 & 0.8369 & 0.7662 & 0.8000 & \textbf{0.8112}\\
SRPN & ResNet-50 & $\mathcal{L}_{triplet}$  & 5426 & 742 & 1271 & 0.8797 & 0.8102 & 0.8435 & 0.7976\\ \hline
RPN & ResNet-101 & -  & 4961 & 756 & 1736 & 0.8678 & 0.7408 & 0.7992 & 0.7470\\
SRPN & ResNet-101 & $\mathcal{L}_{pair}$  & 5136 & 1029 & 1561 & 0.8331 & 0.7669 & 0.7986 & 0.7964\\
SRPN & ResNet-101 & $\mathcal{L}_{triplet}$  & 5399 & 695 & 1298 & 0.8860 & 0.8062 & 0.8442 & 0.7785\\ \hline
RPN & ResNeXt-101 & -  & 4398 & 1501 & 2299 & 0.7456 & 0.6567 & 0.6983 & 0.7678\\
SRPN & ResNeXt-101 & $\mathcal{L}_{pair}$  & 5184 & 994 & 1513 & 0.8391 & 0.7741 & 0.8053 & 0.8024\\
SRPN & ResNeXt-101 & $\mathcal{L}_{triplet}$  & 5482 & 663 & 1215 & \textbf{0.8921} & \textbf{0.8186} & \textbf{0.8538} & 0.7898\\ \Xhline{3\arrayrulewidth}
\end{tabular}
\end{threeparttable}
\end{table*}

\subsubsection{Ablation study}
To select suitable margins for the embedding loss functions as given in Eqs. (\ref{eq: contrastive}) and (\ref{eq: triplet}), we train models with the pair and triplet loss functions by varying the margin from 0.5 to 2.0 respectively. Fig. \ref{fig:sel_m} shows model performance against different margins on the MoNuSeg testing dataset. It is observed that margin $m=1.0$ for the pair loss and $m=2.0$ for the triplet loss yield the best $F1$-scores in the respective cases.

\begin{figure*}
\centering
\includegraphics[width=0.85\textwidth]{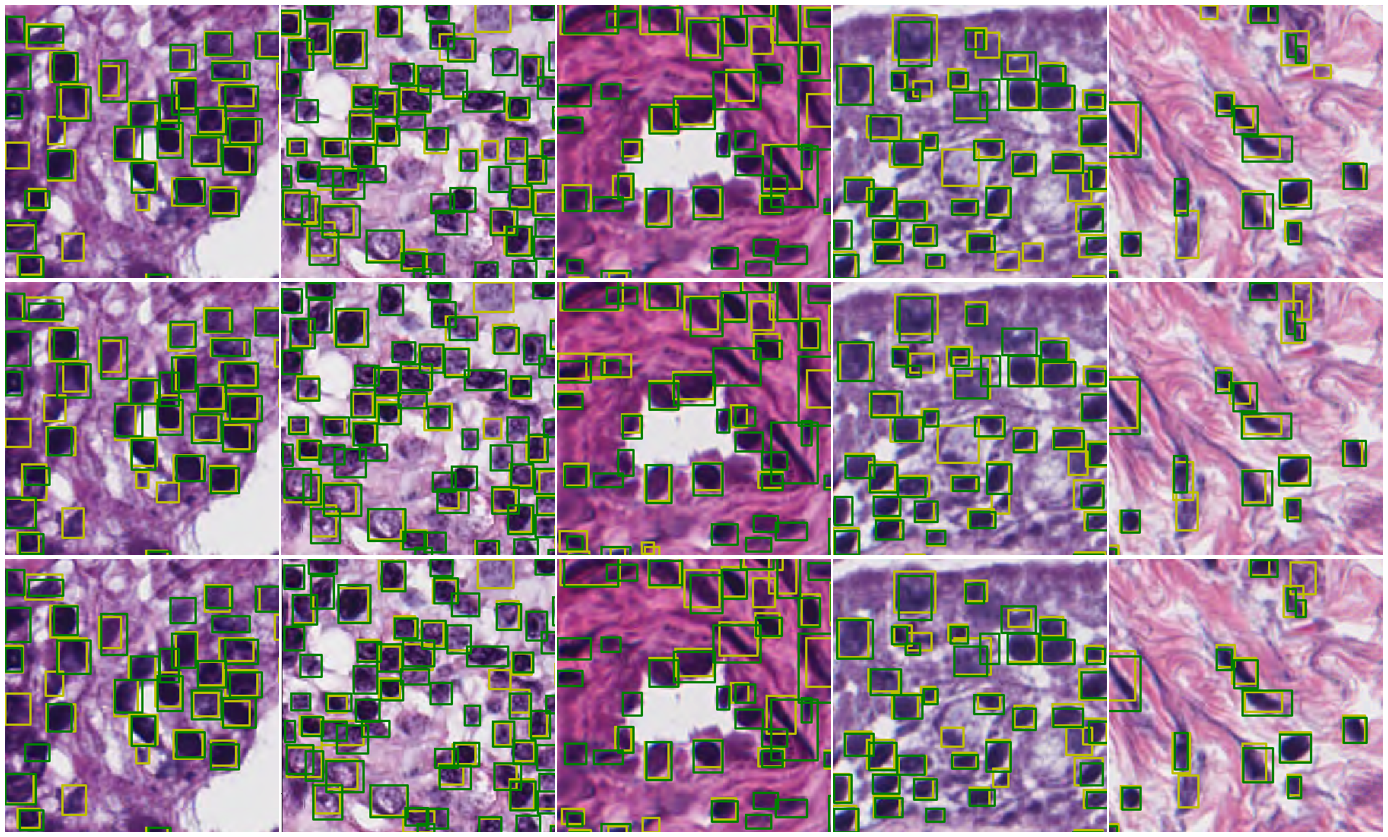}
\caption{\label{fig:monu} Detection of nuclei by models trained with different loss functions. First row: trained without embedding loss; second row: trained with $\mathcal{L}_{pair}$; third row: trained with $\mathcal{L}_{triplet}$. Ground truth is depicted in green bounding boxes and detection in yellow bounding boxes (best view in color).}
\end{figure*}

To investigate the impact of similarity learning on nuclei detection, we test a variety of the proposed models with different CNN backbones (ResNet-50, ResNet-101 or ResNeXt-101) and embedding loss functions (pair or triplet loss). Fig. \ref{fig:monu} shows some samples of ground truth and corresponding detection for a visual assessment.

\begin{figure*}
\centering
\includegraphics[width=0.89\textwidth]{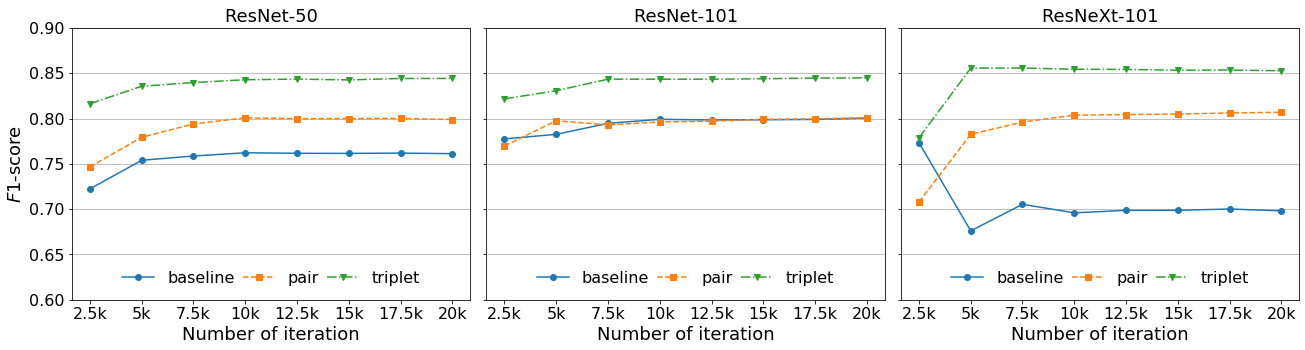}
\caption{\label{fig:f1} $F1$-score of different models with different CNN backbones and embedding losses evaluated on the MoNuSeg testing dataset.}
\end{figure*}

Table \ref{tab:monu_result_archi} presents a comparison among the tested models in different settings. As it can be observed, the performance of these models with embedding loss functions show great improvements in both $F1$-score and average precision compared to the associated baselines (models without embedding loss functions), demonstrating the effectiveness of similarity learning for nuclei detection. For example, the original model with ResNet-50 backbone achieves a $F1$-score of 0.7618, which is significantly increased to 0.8435 when the triplet loss is applied to it. Similar increases of $F1$-scores can also be seen for the other two backbones. Focusing on the $AP$ column, we can see that the models with embedding losses applied, especially the pair loss, outperform their corresponding baselines with clear advantages. Overall, these results evidence a strong positive influence of similarity learning on the performance of nuclei detection models.

To further investigate and compare the performance of each model, we plot $F1$-score against the number of training iteration at a sampling interval of 2500, as depicted in Fig. \ref{fig:f1}. The lines of the triplet models are superior to those of the baseline models in all cases. This further reveals the fact that the employment of similarity learning introduces evident enhancement to the detection performance, due to its excellent ability to distinguish nucleus out of the background. Moreover, the convergence performance of the baseline models changes as the network depth and complexity increase from ResNet-50 to ResNeXt-101 while those of similarity learning models keep relatively stable. After using ResNet-101, the baseline model's $F1$-score is close to that of the network with pair loss, but still far behind that of the triple loss model. However, the training of the baseline model fails with further increase of complexity of network. These observations further validate the efficacy of the embedding losses in leveraging the similarity metric for nuclei detection.

\subsubsection{Comparison against the state-of-the-art}
To demonstrate the effectiveness of the proposed SRPN method for nuclei detection, we compare the performance of different SoTA methods evaluated on the MoNuSeg testing dataset. Table \ref{tab:result_monuseg} shows a comparison of these methods on seen-organ and unseen-organ images. All the deep learning based methods (from method No. 2 to 7) outperform the conventional watershed method (Method 1) on both seen-organ and unseen-organ images by a large margin. Among them, it is observed that the proposed method SRPN achieves the state-of-the-art results. The $F1$-score on seen organs is 3\% higher than the best published method \cite{chen2017dcan} and that of the unseen organ is almost the same with the best result \cite{zhou2019cia}. 

\begin{table}[]
\centering
\begin{threeparttable}
\caption{Performance comparison of different methods for seen-organ and
unseen-organ images.}
\label{tab:result_monuseg}
\begin{tabular}{l|l|l|l}
\Xhline{3\arrayrulewidth}
\multirow{2}{*}{No.} &\multirow{2}{*}{Method} & \multicolumn{2}{c}{$F1$-score} \\ \cline{3-4} 
                        & & Seen         & Unseen         \\ \Xhline{2\arrayrulewidth}
1& Fiji \cite{schindelin2012fiji}                    & 0.6402        & 0.6978        \\ \hline
2& CNN3 \cite{kumar2017dataset}                    & 0.8226        & 0.8322        \\ \hline
3& DCAN \cite{chen2017dcan}                   & 0.8265        & 0.8214        \\ \hline
4& PA-Net \cite{liu2018path}                  & 0.8156        & 0.8336        \\ \hline
5& BES-Net \cite{oda2018besnet}                 & 0.8118        & 0.7952        \\ \hline
6& CIA-Net \cite{zhou2019cia}                 & 0.8244        & \textbf{0.8458}       \\ \hline
7& \textbf{Proposed SRPN}                & \textbf{0.8579}            & 0.8427              \\ \Xhline{3\arrayrulewidth}
\end{tabular}
\end{threeparttable}
\end{table}

In the nuclei detection task, similarity learning can significantly enhance the classification ability of the proposed model, especially when the triple loss is used. We argue that this method's ability in producing better results on data with low diversity makes it particularly suitable for tasks on cell-level object detection in histology images where traditional detection methods cannot obtain better performance. 
It can extract more discriminative features for nuclei detection and maintain its performance in spite of the change of CNN architectures. The proposed SRPN model shows superior performance on nuclei detection in both seen and unseen organ images, because its ability in leveraging the similarity metric for instance classification.

\subsection{Evaluation on signet ring cells detection}
\label{sec: exp}
We further test the proposed method on a much more challenging task - signet ring cell detection. Signet ring cell is a type of abnormal cell that is most frequently associated with stomach cancer. A tumour is defined as signet ring cell adenocarcinoma when it is composed of at least 50\% signet ring cells. Generally, a tumour has a worse prognosis when a significant number of signet ring cells present. It is of clinical importance to detect and count the number of signet ring cells in a region. This can be used as valuable clues to help pathologists understand and evaluate the degree of tissue lesion. However, due to its large morphological variations and other complexities, signet ring cell detection remains a challenging task. Therefore in this section, we validate the performance of the proposed SRPN method by applying it to solve the task of signet ring cell detection.
\subsubsection{Dataset}
The used dataset is released for the Digestive-System Pathological Detection and Segmentation Challenge\footnote{\href{https://digestpath2019.grand-challenge.org/Dataset/}{https://digestpath2019.grand-challenge.org/Dataset/}}. There are in total 77 histology images with annotations from 20 patients. All the images are acquired from either gastric mucosa or intestine. The average size of each image is about $2000 \times 2000$ pixels and there is a total of 9,710 signet ring cells annotated by experienced pathologists in the format of bounding boxes. To validate the networks, we randomly split the images into a training set and a validation set with a ratio of $4:1$. Images from the training set are then cropped into small patches for training. The size of patches is defined as $600 \times 600$ pixels here according to the network input requirement.

\subsubsection{Evaluation criteria}
In the annotated images, pathologists can only guarantee that the labeled cells all belong to the signet ring cell category, but cannot exhaustively label all the signet ring cells presented in the images, especially in the overcrowded regions. In this situation, it is not possible to use average precision or $F1$-score to validate and compare the detection performance. Evaluation metrics here include 1) recall $R=\frac{TP}{TP+FN}$,
and 2) score of normal region false positives $S_{nr}= \frac{max(100-FP_{nr}, 0)}{100}$,
where $FP_{nr}$ is mean normal region false positives counted on a set of extra negative images. This set contains 378 extra images extracted from normal regions, each of size $2000 \times 2000$ pixels, and are employed only for evaluation purposes. Since there is no signet ring cells present in the negative images, all the predicted bounding boxes in the negative images are added to $FP_{nr}$. 
The evaluation results are compared against baseline networks without the embedding losses, but no comparison with external methods can be presented in this paper as the results are not yet published to date.

\subsubsection{Results and discussion}
\label{sec: results}
We comprehensively test and compare different models (Faster R-CNN and RetinaNet) with different CNN backbones and/or embedding loss functions on both the validation set and the extra negative image set. Table \ref{tab:result_ringcell} lists the recall value $R$ and the score on normal regions $S_{nr}$ at a confidence threshold of 0.5 of each model. From group 1 to 3 for different models, the same observation can be drawn that models with embedding loss functions outperform those baselines in terms of the score on normal region $S_{nr}$. However, the proposed similarity learning models achieve slightly lower recall $R$ values in most cases. In particular, it can be observed from group 1 that the SRPN model with triplet losses achieves superior average performance when compared to the corresponding RPN model without embedding losses due to the improvement on $S_{nr}$. 

\begin{table*}[]
\centering
\begin{threeparttable}
\caption[Comparison of signet ring cell detection with different models, CNN backbones and embedding loss functions.]{Comparison of signet ring cell detection results of different models, CNN backbones and embedding loss functions. $R$@0.5: recall at a confidence threshold of 0.5; $S_{nr}$@0.5: score on normal regions at a confidence threshold of 0.5.}
\label{tab:result_ringcell}
\begin{tabular}{c|l|l|l|l|c}
\Xhline{3\arrayrulewidth}
Group & Model & Backbone & $\mathcal{L}_{embed}$ & R@0.5 & $S_{nr}$@0.5  \\
\Xhline{2\arrayrulewidth}
\multirow{3}{*}{1} & RPN & ResNet-50 & - & 0.8428 & 0.07  \\
 & SRPN & ResNet-50 & $\mathcal{L}_{pair}$ & \textbf{0.8482} & 0.00  \\
 & SRPN & ResNet-50 & $\mathcal{L}_{triplet}$ & 0.7796 & 0.39  \\  \hline
\multirow{3}{*}{2} & Faster R-CNN & ResNet-50 & - & 0.6141 & 0.71  \\
 & Faster R-CNN & ResNet-50 & $\mathcal{L}_{pair}$ & 0.6312 & 0.68  \\
 & Faster R-CNN & ResNet-50 & $\mathcal{L}_{triplet}$ & 0.6004 & 0.78  \\ \hline
\multirow{3}{*}{3} & RetinaNet & ResNet-50 & - & 0.6273 & 0.72  \\
 & RetinaNet & ResNet-50 & $\mathcal{L}_{pair}$ & 0.57 & 0.83  \\
 & RetinaNet & ResNet-50 & $\mathcal{L}_{triplet}$ & 0.5602 & 0.83  \\ \hline \hline
\multirow{3}{*}{4} & RetinaNet & ResNet-50 & - & 0.6273 & 0.72  \\
 & RetinaNet & ResNet-101 & - & 0.5999 & 0.77  \\
 & RetinaNet & ResNeXt-101 & - & 0.6009 & 0.78  \\ \hline
\multirow{3}{*}{5} & RetinaNet & ResNet-50 & $\mathcal{L}_{pair}$ & 0.57 & 0.83  \\
 & RetinaNet & ResNet-101 & $\mathcal{L}_{pair}$ & 0.547 & 0.85 \\
 & RetinaNet & ResNeXt-101 & $\mathcal{L}_{pair}$ & 0.5225 & \textbf{0.88}  \\\hline
\multirow{3}{*}{6} & RetinaNet & ResNet-50 & $\mathcal{L}_{triplet}$ & 0.5602 & 0.83  \\
 & RetinaNet & ResNet-101 & $\mathcal{L}_{triplet}$ & 0.5235 & 0.86  \\
 & RetinaNet & ResNeXt-101 & $\mathcal{L}_{triplet}$ & 0.5215 & 0.85 \\
\Xhline{3\arrayrulewidth}
\end{tabular}
\end{threeparttable}
\end{table*}

\begin{figure*}
\centering
\includegraphics[width=0.95\textwidth]{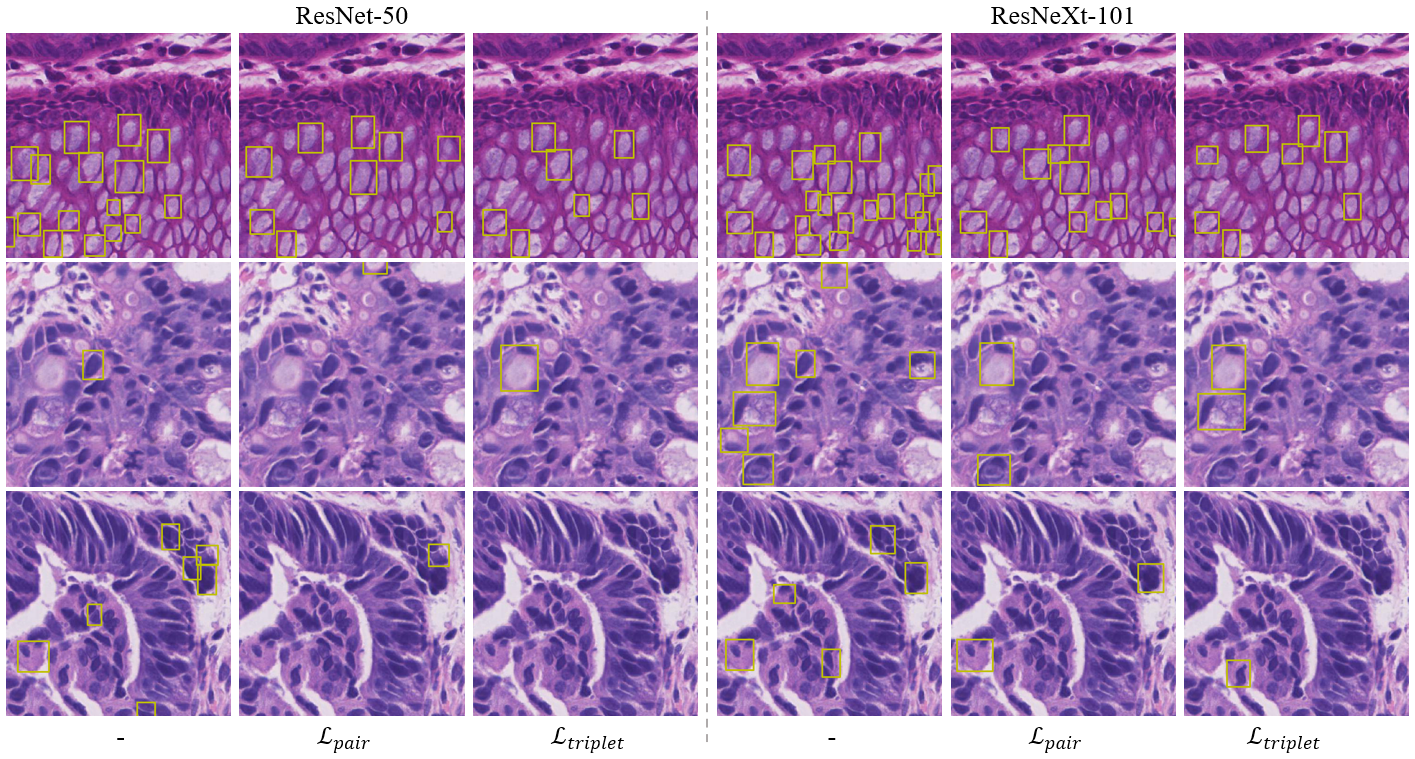}
\caption{\label{fig:ringcell} False positives on negative images of normal region detected by models with different CNN backbones and embedding loss functions (best view in color).}
\end{figure*}

\begin{figure*}
\centering
\includegraphics[width=0.75\textwidth]{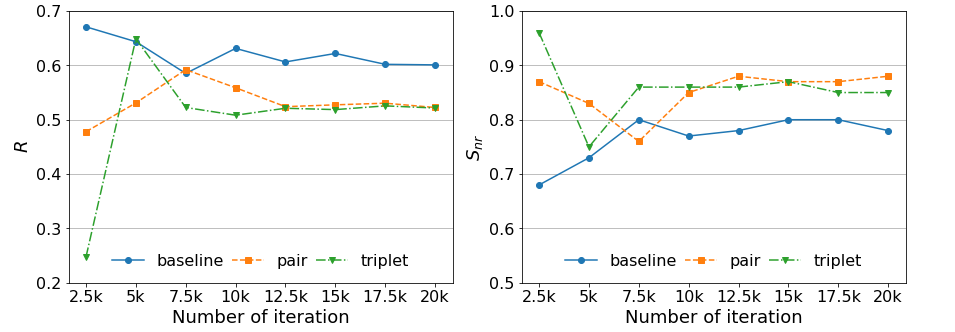}
\caption{\label{fig:ringcell_per} Performance comparison among models trained with/without different embedding loss functions, evaluated on the validation set of signet ring cell detection.}
\end{figure*}

To investigate the impact of the CNN architectures on signet ring cells detection, we test the RetinaNet model with multiple CNN backbones including ResNet-50, ResNet-101 and ResNeXt-101. As can it be seen from group 4 to 6 of Table \ref{tab:result_ringcell}, the light-weight architecture ResNet-50 perform better on recall $R$ while deeper and more complex architectures ResNet-101 and ResNeXt-101 perform better on $S_{nr}$, resulting in a close average performance of different backbones.

To observe changes of performance against training iteration and to compare the performance of different embedding loss functions, we plot the performance of RetinaNet with different embedding loss functions, on the signet ring cell validation set (Fig. \ref{fig:ringcell_per}). As presented in the figure, normal region scores $S_{nr}$ of models applied embedding loss are better than those of the baselines while the recall $R$ values demonstrate in a different way. The observation that models with embedding losses outperform baselines on normal region scores $S_{nr}$ can also be validated in Fig. \ref{fig:ringcell}, which shows false positives of signet ring cell on negative images of normal region predicted by different models. As can it be seen from the figure, the numbers of false positives decrease significantly when embedding loss functions are applied.

These result together show that the proposed SRPN networks with a similarity learning scheme present excellent ability in discriminating true and untrue instances, and can thus avoid challenging false positives being detected while maintaining a high true positive rate.

\section{Conclusion}
We present a similarity based region proposal network (SRPN) to accurately detect nuclei and cells in histology images. This challenging cell-level object detection problem is formulated as a multi-task learning process, namely, instance localisation and classification. A similarity metric is used to improve classification performance. To apply similarity leaning, we introduce an embedding layer to the SRPN architecture for building networks, which allows us to train networks with embedding loss functions. Networks trained with embedding losses are able to learn discriminative features based on the similarities and use them for instance classification. The proposed SRPN has been evaluated on two cell-level object detection benchmarks. Significant improvement are introduced by exploiting embedding losses, demonstrating the effectiveness of the similarity learning approach for nuclei and cells detection. Specifically, experimental results show that SRPN yield outstanding performance on the MoNuSeg benchmarks for nuclei detection compared to previous methods, and on the signet ring cell detection benchmark when compared against baseline networks.

\bibliographystyle{IEEEbib}
\bibliography{ring_cell_detection}

\end{document}